\documentclass{article}
\usepackage{spconf,amsmath,graphicx}
\usepackage{newtxmath}
\usepackage{booktabs} 

\title{SynGen: A Syntactic Plug-and-play Module for Generative \\ Aspect-based Sentiment Analysis}
\name{Chengze Yu\textsuperscript{*}, 
        Taiqiang Wu\textsuperscript{*}\thanks{* Equal contribution.},
        Jiayi Li,
        Xingyu Bai,
        Yujiu Yang$^\dagger$\thanks{$^\dagger$ Corresponding author.}}
        
\address{Tsinghua Shenzhen International Graduate School, Tsinghua University\\
    \{ycz21, wtq20, lijy20, bxy20\}@mails.tsinghua.edu.cn, yang.yujiu@sz.tsinghua.edu.cn}

\usepackage{bibentry}
\begin{document}
%
\maketitle

\begin{abstract}
Aspect-based Sentiment Analysis~(ABSA) is a sentiment analysis task at fine-grained level.
Recently, generative frameworks have attracted increasing attention in ABSA due to their ability to unify subtasks and their continuity to upstream pre-training tasks.
However, these generative models suffer from the neighboring dependency problem that induces neighboring words to get higher attention.
In this paper, we propose SynGen, a plug-and-play syntactic information aware module.
As a plug-in module, our SynGen can be easily applied to any generative framework backbones.
The key insight of our module is to add syntactic inductive bias to attention assignment and thus direct attention to the correct target words.
To the best of our knowledge, we are the \textbf{first one} to introduce syntactic information to \textbf{generative} ABSA frameworks.
Our module design is based on two main principles: (1) maintaining the structural integrity of backbone PLMs and (2) disentangling the added syntactic information and original semantic information.
Empirical results on four popular ABSA datasets demonstrate that SynGen enhanced model achieves a comparable performance to the state-of-the-art model with relaxed labeling specification and less training consumption.
\end{abstract}


\section{Introduction}
Aspect-based Sentiment Analysis~(ABSA)~\cite{14res_14lap} is a fine-grained sentiment analysis task.
A typical ABSA task includes three components: aspects, sentiment polarities and opinions.
The goal of ABSA is to classify sentiment polarity around the aspect based on the corresponding opinion.
For example, in the sentence "\textit{The food is good but the service is bad}", the aspects are \textit{food} and \textit{service}, while the relevant opinions are \textit{good} and \textit{bad}, respectively.
According to the opinions, the sentiment polarity of the aspect \textit{food} is positive, and the sentiment polarity of \textit{service} is negative.
Recently, generative frameworks have been widely adopted for ABSA~\cite{unified_generative, SentiPrompt, generative1}.
Most of them are based on encoder-decoder structure Pre-trained Language Models~(PLMs) and view ABSA as a text generation task.
These generative frameworks achieve promising performance due to their ability of unifying ABSA subtasks and their continuity to upstream pre-training tasks.

\begin{figure}[t]
    \centering
    \includegraphics[width=\linewidth]{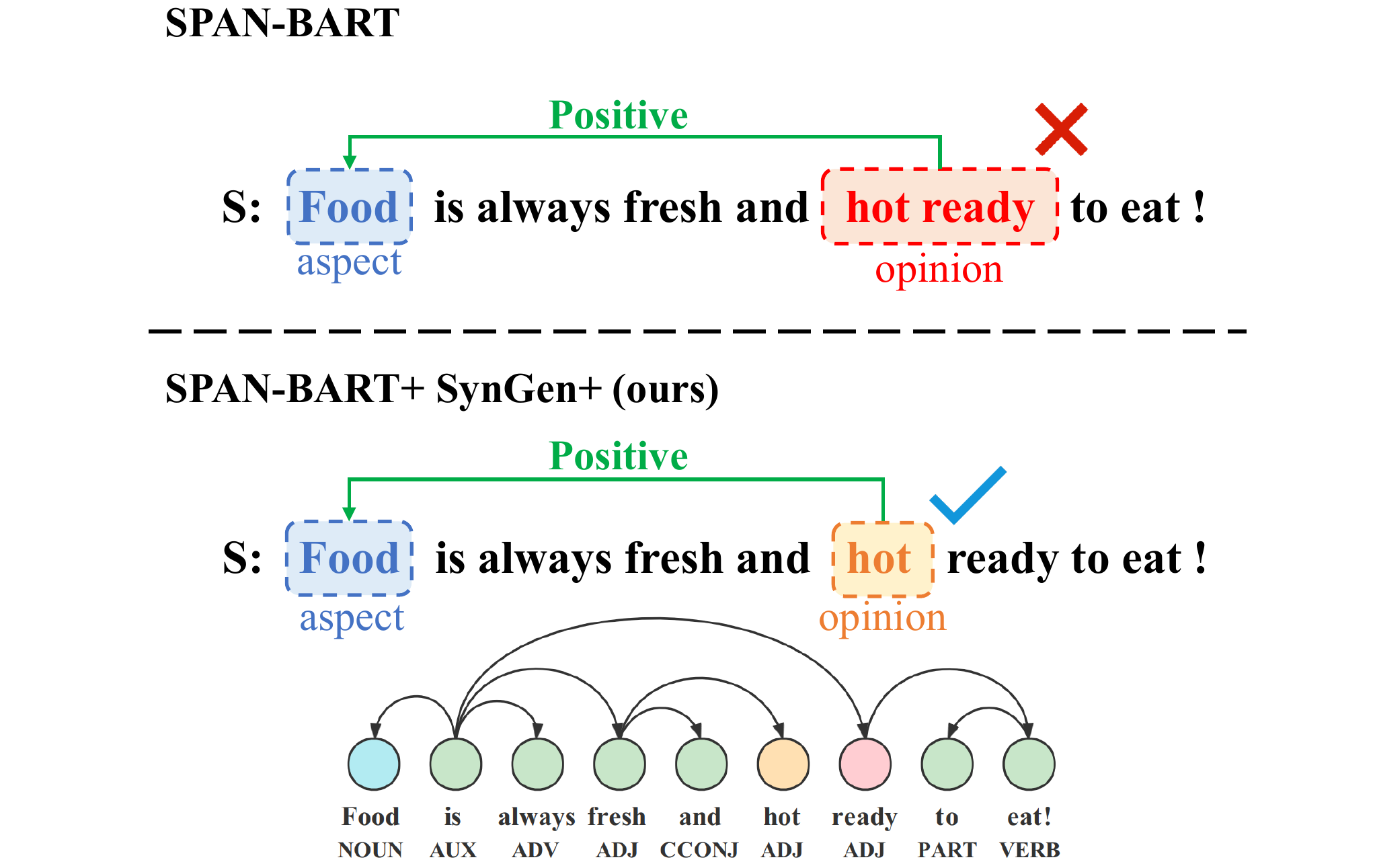}
    \caption{The output for baseline SPAN-BART~\cite{unified_generative} and our SynGen under the same input sentence.
    The generative framework SPAN-BART extracts a redundant word \textit{ready} and our SynGen avoids this mistake by employing the syntactic information.}
    \vspace{-1em}
\end{figure}

However, generative frameworks suffer from neighboring dependency problem~\cite{neighboring, neighboring2} that induces neighboring words to get higher attention.
As illustrated in Figure 1, given the sentence "\textit{Food is always fresh and hot ready to eat!}", the target ABSA output should be (Food, hot, positive). 
But the baseline SPAN-BART~\cite{unified_generative} extracts a wrong triplet (Food, hot ready, positive) with a redundant word \textit{ready} in the opinion.
Since the words \textit{hot} and \textit{ready} are adjacents, the SPAN-BART improperly treats them as a unit.


To address this issue, we propose SynGen, a plug-and-play syntactic information aware module.
The key insight is to introduce syntactic inductive bias to attention assignment which will guide words to focus more on other related words.
In practice, we employ a syntactic dependency tree to model the syntactic information.
As shown in the lower portion of Figure 1, the adjacent words \textit{hot} and \textit{ready} are far apart from each other in the view of a syntactic dependency tree.
Being aware of this prior distance knowledge , the model can distinguish these words from a unit.

To properly adapt our module to the generative frameworks, we design the SynGen following two main principles: (1) maintaining the \textbf{structural integrity} of backbone PLMs and (2) \textbf{disentangling} the added syntactic information and original semantic information.

The goal of maintaining \textbf{structural integrity} is to narrow the gap with upstream PLMs.
Since the PLM is sophisticated enough, any slightest unreasonable change will compromise the model performance.
For example, modeling syntactic information by adding the word embeddings with syntactic information embeddings will bring in non-pretrained input for the PLMs and thus hurt the semantic capturing performance.
In this case, we implement our module by appending an independent syntactic channel parallel working with the PLM backbone.
We view the original encoder as a semantic channel and fuse it with a syntactic channel to get more comprehensive representations.
Moreover, we design a dynamic gate mechanism to point-wisely add the outputs from these two channels.

The goal of \textbf{disentangling} the appendant syntactic information and original semantic information is to avoid error accumulation.
Following previous studies~\cite{graph2, graph3, graph4, taki, graph1}, we build a syntactic dependency tree based graph and use Graph Neural Networks~(GNN) to model the syntactic information.
The syntactic channel should exclude the influence of biased PLM backbone output which is the reason of neighboring dependency problem~\cite{neighboring}.
Hence, we initialize the graph nodes with a sequence of part-of-speech~(POS) tags rather than directly using the backbone encoder output.
Moreover, owing to the parallel design and disentangling strategy, our SynGen is a plug-and-play module which can be easily applied to any other generative frameworks.

In summary, our main contributions are as follows:

$\bullet$ We propose a plug-and-play module SynGen to introduce syntactic information to generative ABSA frameworks. 
To the best of our knowledge, we are the first one to add syntactic information to generative frameworks.

$\bullet$ Our SynGen can solve the neighboring dependency problem by guiding the attention from aspect to corresponding opinion.

$\bullet$ We implement experiments on popular ABSA datasets. 
The experimental results demonstrate that our module achieves a promising improvement compared to strong baselines.

\section{Methodology}

\begin{figure*}[!h]
	\centering
	\includegraphics[width=0.75\linewidth]{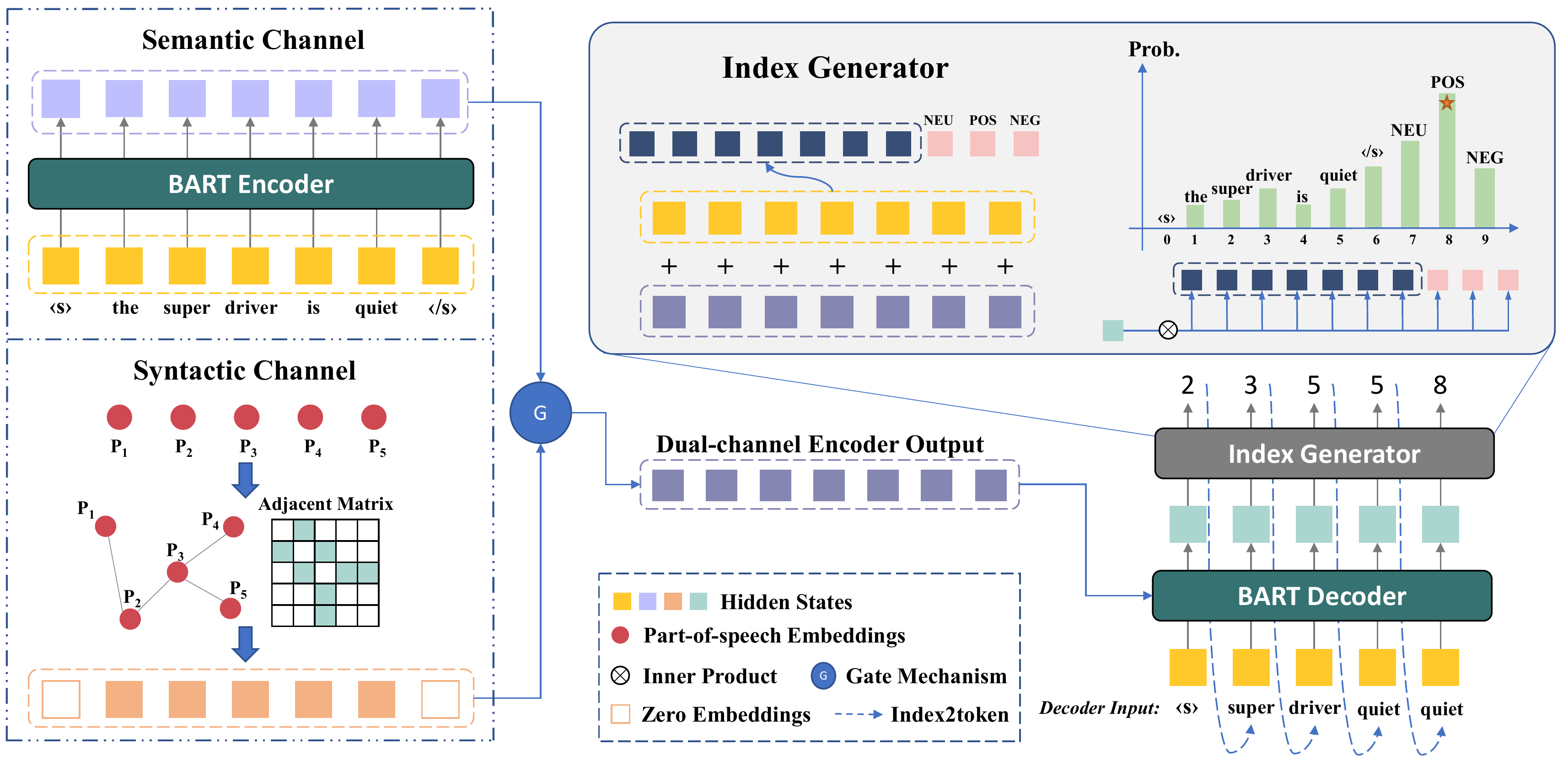}
	\caption{
	An overview of the SynGen enhanced model.
	The model contains a dual-channel encoder and a pointer network decoder.
	The dual-channel encoder models syntactic and semantic information in parallel and their outputs are fused by a dynamic gate mechanism.
	The decoder sequentially generates target predictions from the candidate indexes.
	}
	\label{framework}
	\vspace{-1em}
\end{figure*}

\subsection{Task Formulation}
In this paper, we apply our module to three ABSA subtasks: Aspect Term Extraction and Sentiment Classification (AESC), Pair Extraction and Triplet Extraction.
Given an input sentence $X=[x_1, x_2, ..., x_n]$, the target output is a series of predictions $Y=[p_1, p_2,$ $ ..., p_k]$, where $n$ and $k$ represent the number of tokens and predictions, respectively.
In each prediction, we use $a$ and $o$ to represent the aspect and opinion, use superscripts $\cdot^s$ and $\cdot^e$ to indicate the start and end of index, and use $s\in\{neutral, positive, negative\}$ to denote the sentiment polarity.

The prediction of three subtasks is as follows:

$\bullet$ AESC: $p_i=[{a_i}^s, {a_i}^e, s_i]$,

$\bullet$ Pair Extraction: $p_i=[{a_i}^s, {a_i}^e, {o_i}^s, {o_i}^e]$,

$\bullet$ Triplet Extraction: $p_i=[{a_i}^s, {a_i}^e,$ ${o_i}^s, {o_i}^e, s_i]$.

\subsection{Model Architecture}
We choose SPAN-BART~\cite{unified_generative} as our backbone and enhance it with the plug-in module SynGen.
The overview of the enhanced model is shown in Figure 2.

\noindent\textbf{Semantic Channel.}
In the \textbf{semantic channel}, we employ BART~\cite{BART}  to model input sentence $X$ into hidden states $H^{se}$.
Two special tokens $\langle s \rangle$ and $\langle /s \rangle$ represent the beginning and ending, respectively.
\begin{align}
  \mathbf{E}^{se}& =\mathrm{BARTEmbed}(\mathrm{\langle s \rangle};X;\mathrm{\langle /s \rangle}) \nonumber\\
     & =[e^{se}_{\langle s \rangle}, e^{se}_1, e^{se}_2, ..., e^{se}_n, e^{se}_{\langle /s \rangle}], \\
  \mathbf{H}^{se} & =\mathrm{BARTEnocder}(\mathbf{E}^{se}),
\end{align}
where $\mathbf{E}^{se},\mathbf{H}^{se} \in \mathbb{R}^{(n+2) \times d}$, superscript $\cdot^{se}$ denote the semantic channel and $d$ is the hidden state dimension.

\noindent\textbf{Syntactic Channel.}
In the \textbf{syntactic channel}, we build a bidirectional graph $\mathbf{G}$ based on the syntactic dependency tree.
The adjacent matrix $\mathbf{A} \in \mathbb{R}^{n \times n}$ of G  is defined as follows:
\begin{align}
  \mathbf{A}_{ij} = \begin{cases}
    1,&  \text{if}  \  x_i, x_j \  \text{connect in dependency tree}, \\
    0,&  \text{otherwise}.
  \end{cases} 
\end{align}
After building the graph, we use a sequence of POS tags $P=[pos_1, pos_2, ..., pos_n]$ to initialize the graph nodes.
Each POS tag $pos_i$ is derived from its corresponding word\footnote{We use the spaCy toolkit to obtain the dependency tree and part-of-speech~(POS) tags: \url{https://spacy.io/}.}.
We use a learnable embedding layer $\mathrm{POSEmbed}$ to embed the POS tags as follows:
\begin{align}
  \mathbf{E}^{sy}=\mathrm{POSEmbed}(P)=[e^{sy}_1, e^{sy}_2, ..., e^{sy}_n],
\end{align}
where $\mathbf{E}^{sy} \in \mathbb{R}^{n \times d}$ and superscript $\cdot^{sy}$ represents the syntactic channel.
Then we initialize the graph nodes by the above POS embeddings as $h^0_i=e^{sy}_i$.

After that, we employ the Graph Attention Networks (GAT)~\cite{GAT} to update nodes embeddings:
\begin{align}
  & h^{l+1}_i = \sum_{j\in \mathcal{N}(i)}\alpha^l_{i,j}h^l_j W^l, \\
  & \alpha^l_{i,j} = \mathrm{softmax}_i(e^l_{i,j}), \\
  & e^l_{i,j} = \mathrm{LeakyReLU}([h_i W^l||h_j W^l] \overrightarrow{a}^T),
\end{align}
where $h^l_i \in \mathbb{R}^{1 \times d}$ is the hidden state of the $i$-th node in the $l$-th layer.
$W^l \in \mathbb{R}^{d \times d}$ and $\overrightarrow{a} \in \mathbb{R}^{1 \times 2d}$ denotes the learnable weight matrices.
In this study, the number of GAT layers $L$ is set to 2.
We choose the last layer hidden states $\mathbf{H}^L$ as the syntactic channel output $\mathbf{H}^{sy}$.
Moreover, we pad the output with zero embeddings to align with the special tokens $\langle s \rangle$ and $\langle /s \rangle$:
$\mathbf{H}^{sy} = [\mathbf{0}; \mathbf{H}^{L}; \mathbf{0}]$.

\noindent\textbf{Gate Mechanism.}
To properly fuse the two channel outputs, we design a \textbf{gate mechanism} which dynamically conditions the syntactic weight according to the semantic channel output:
\begin{align}
    \mathbf{H}^{e} = \mathbf{H}^{se}+\sigma\{\mathrm{Linear}(\mathbf{H}^{se})\} \odot \mathbf{H}^{sy}.
\end{align}
$\mathrm{Linear}$ denotes a linear layer reshaping the vector into one dimension and $\odot$ represents the Hadamard product. 
For the activation function $\sigma$, we use the sigmoid function. 

\noindent\textbf{Decoder.}
As for the \textbf{decoder}, we adopt point networks which generate the word indexes $Y_{<t}=[y_1, y_2, ..., y_{t-1}]$ sequentially.
An Index-to-Token module is applied to convert $Y_{<t}$ to the appropriate hidden states:
\begin{align}
  \hat{y}_t = \begin{cases}
    X_{y_t},&  \text{if}  \  y_t \  \text{is a pointer index},\\
    C_{y_t - n},&  \text{if} \  y_t \  \text{is a polarity index},
  \end{cases}
\end{align}
where $C=\{neutral, positive, negative\}$ is the sentiment polarity list.
After converting the previous decoder outputs $\hat{Y}_{<t}$, we use the BART decoder to get the $t$-step hidden state:
\begin{align}
  \mathbf{h}_t^d & = \mathrm{BARTDecoder}(\mathbf{H}^e; \hat{Y}_{<t}),
\end{align}
where the superscript $\cdot^d$ represents the decoder output and the subscript $t$ denotes the t-th step.

To predict the $y_t$ probability distribution, we concatenate $\mathbf{\bar{H}^e}$ and $\mathbf{C}^d$ and make inner product with $\mathbf{h}_t^d$:
\begin{align}
  & \mathbf{\hat{H}}^e = \mathrm{MLP}(\mathbf{H}^e), \\
  & \mathbf{\bar{H}}^e = \alpha \mathbf{\hat{H}}^e + (1-\alpha) \mathbf{E}^{se}, \\
  & \mathbf{C}^d = \mathrm{BARTEmbed}(C), \\
  & \mathrm{Pro}_t = \mathrm{softmax}([\mathbf{\bar{H}^e};\mathbf{C}^d]  \mathbf{h}_t^d),
\end{align}
where ; denotes concatenation, $\mathbf{E}^e,\mathbf{H}^e,\mathbf{\hat{H}}^e,\mathbf{\bar{H}}^e \in \mathbb{R}^{(n+2) \times d}$,  $\mathbf{C}^d \in \mathbb{R}^{3 \times d}$, $\alpha$ represents the proportionality coefficient and $\mathrm{Pro}_t \in \mathbb{R}^{(n+5)}$ is the predicted distribution of $y_t$ among candidate indices.

In the training phase, we use the negative log-likelihood as the loss function and adopt the teacher forcing method to optimize our model.
Moreover, in the inference phase, we use the beam search to generate the prediction result.


\begin{table*}[]
\resizebox{1.8\columnwidth}{!}
{
\begin{tabular}{lccc|ccc|ccc|ccc}
\hline
Model & \multicolumn{3}{c|}{14res}                       & \multicolumn{3}{c|}{14lap}                       & \multicolumn{3}{c|}{15res}                       & \multicolumn{3}{c}{16res}                        \\ \cline{2-13} 
                       & AESC           & Pair           & Triplet        & AESC           & Pair           & Triplet        & AESC           & Pair           & Triplet        & AESC           & Pair           & Triplet        \\ \hline
CMLA+                   & 70.62          & 48.95          & 43.12          & 56.90          & 44.10          & 32.90          & 53.60          & 44.60          & 35.90          & 61.20          & 50.00          & 41.60          \\
RINANTE+                & 48.15          & 46.29          & 34.03          & 36.70          & 29.70          & 20.00          & 41.30          & 35.40          & 28.00          & 42.10          & 30.70          & 23.30          \\
Li-unified+             & 73.79          & 55.34          & 51.68          & 63.38          & 52.56          & 42.47          & 64.95          & 56.85          & 46.69          & 70.20          & 53.75          & 44.51          \\
Peng-two-stage         & 74.19          & 56.10          & 51.89          & 62.34          & 53.85          & 43.50          & 65.79          & 56.23          & 46.79          & 71.73          & 60.04          & 53.62          \\
JET-BERT               & --             & --             & 63.92          & --             & --             & 50.00          & --             & --             & 54.67          & --             & --             & 62.98          \\
Dual-MRC               & 76.57          & 74.93          & 70.32          & 64.59          & 63.37          & 55.58          & 65.14          & 64.97          & 57.21          & 70.84          & 75.71          & 67.40          \\
SPAN-BART              & 78.47          & 77.68          & 72.46          & 68.17          & 66.11          & 57.59          & 69.95          & 67.98          & 60.11          & 75.69          & 77.38          & 69.98          \\
SyMux                  & 78.68          & \textbf{79.42} & \textbf{74.84} & \textbf{70.32} & 67.64          & 60.11          & 69.08          & \textbf{69.82} & 63.13          & \textbf{77.95} & \textbf{78.82} & \textbf{72.76} \\ \hline
Ours                   & \textbf{79.72} & 77.59          & 74.02          & 70.06          & \textbf{68.53} & \textbf{60.71} & \textbf{71.61} & 69.35          & \textbf{64.06} & 77.51          & 77.34          & 71.26          \\
\enspace w/o graph              & 77.54          & 76.40          & 71.92          & 70.05          & 67.85          & 60.29          & 70.50          & 68.92          & 62.86          & 73.04          & 75.84          & 69.13          \\
\enspace w/o gate               & 77.37          & 76.03          & 71.90          & 69.49          & 68.09          & 60.43          & 71.45          & 68.68          & 63.11          & 75.33          & 76.16          & 70.14          \\
\enspace w/o graph\&gate        & 78.72          & 77.09          & 73.36          & 68.21          & 65.84          & 58.17          & 65.80          & 63.44          & 57.44          & 74.84          & 75.06          & 68.82          \\ \hline
\end{tabular}
}
\centering
\caption{Comparison F1 scores for AESC, Pair and Triplet. The best results are highlighted in bold. The suffix “+” denotes being modified by Peng~\cite{penga} for being capable of AESC, Pair and Triplet.}
\vspace{-1em}
\end{table*}

\section{Experiments}

\subsection{Experiment Setting}
\textbf{Datasets.}
We evaluate our method on four popular ABSA datasets, including Rest14, Laptop14, Rest15 and Rest16.
These original datasets are proposed by the SemEval Challenges~\cite{14res_14lap, 15res, 16res}.
To obtain high quality triplets, we adopt a refined version provided by Haiyun Peng~\cite{penga} which distills the triplets based on two previous relabeld versions~\cite{D17, D19}.

\noindent \textbf{Baselines.}
To fairly evaluate the performance of the SynGen enhanced model, we choose the following baselines for three ABSA subtasks:
CMLA~\cite{D17}, RINANTE~\cite{RINANTE}, Li-unified~\cite{Li-unified}, Peng-two-stage~\cite{penga}, JET-BERT~\cite{pengb}, Dual-MRC~\cite{unified1}, SPAN-BART~\cite{unified_generative} and SynMux~\cite{SynMux}.


\noindent \textbf{Implementation.}
In this study, all experiments are implemented on a 24G RTX3090 GPU, taking about an hour to complete 200 epochs of training with a batch size of 48.
We employ the BART as the backbone and set the hidden dimensions to 768.
The GAT learning rate is set to 1e-5 and the other parts learning rate is set to 1e-4.
Following previous studies, we use F1 scores as metrics and a predicted aspect or opinion is judged to be correct only if the predicted span exactly matches the start and end boundaries of the ground truth.

\subsection{Main Results}
As shown in Table 1, SynGen enhanced model achieves a comparable performance to the state-of-the-art~(SOTA) model SyMux but with relaxed labeling specification and less training consumption.
The SOTA model ensembles three version of ABSA datasets to acquire additional labeling for all seven tasks while our SynGen needs no such data augment.
And the SOTA model contains seven decoders to handle each seven ABSA task, which largely increases the training consumption.
Meanwhile, our SynGen enhanced model outperforms the original baseline SPAN-BART for almost every task. 
Especially for Triplet, we get a 3.95\% F1 promotion on 15res, representing a convincing superiority of the method.

\subsection{Analysis}
\textbf{Ablation.}
To further evaluate the necessity of each part in SynGen, we remove the graph and gate mechanism in turn to verify the effectiveness of these two modules.
As shown in the Table 1, removing either graph or gate mechanism leads to a decline on model performance, which demonstrates that both graph and gate mechanism are indispensable.
Besides, removing the dynamic gate mechanism leads to a larger drop than removing the graph.
Such a phenomenon strengthens the importance of the proper algorithm to integrate syntactic and semantic information.

\noindent \textbf{Attention Weight Study.}
Intuitively, our SynGen can provide a syntactic inductive bias to the generative frameworks and thus address the neighboring dependency issue.
To evaluate this assumption, we implement experiments on two version ($\mathrm{Da}$~\cite{penga},  $\mathrm{Db}$~\cite{pengb}) of dataset 14res, and analyze the attention weight gap between SynGen enhanced model and the baseline model.
To fully evaluate the magnitude of attention weight gaps, we evaluate from three perspectives:
1) \textbf{Value} represents the absolute value gap of attention weights. 
2) \textbf{Rank} denotes the changes in attention weight ranks. 
3) \textbf{Prop.} indicates the proportion difference between the two models.




\begin{table}[!t]
\begin{tabular}{l|ccc}
\toprule
\textbf{Dataset} & \textbf{Value} & \textbf{Rank} & \textbf{Prop.} \\ \hline
Da\_14res        & 0.00789        & 0.11065       & 0.36691        \\
Db\_14res        & 0.00155        & 0.28385       & 0.11654        \\ \bottomrule
\end{tabular}
\centering
\caption{An illustration on how SynGen contributes the attention weights from aspect to opinion.}
\vspace{-1 em}
\end{table}

As shown in Table 2, with the enhancement of SynGen, the attention weights from aspect to opinion boost under all three perspectives.
Especially in terms of the proportion, the enhanced model achieves a relative promotion of 36.69\% in version $D_a$ and 11.65\% improvement in version $D_b$.
This demonstrates that our SynGEN successfully guides the attention from aspect to the corresponding opinion and thus solves the neighboring dependency problem.

\section{Conclusion}

In this paper, we propose SynGen, a plug-and-play syntactic information aware module.
Our module effectively introduces syntactic inductive bias on attention assignment and thus solves the neighboring dependency problem.
To the best of our knowledge, we are the first one to introduce syntactic information to generative ABSA frameworks.
Moreover, our module maintains the structural integrity of backbone PLMs by inserting a parallel syntactic channel.
Empirical results demonstrate that the SynGen enhanced model achieves a comparable performance to the state-of-the-art model with relaxed labeling specification and less training consumption.
Further analysis demonstrates that our module can guide the attention from aspect to corresponding opinion.

\clearpage 
\bibliographystyle{IEEEbib}
\bibliography{main}

\begin{thebibliography}{10}

\bibitem{14res_14lap}
Maria Pontiki, Dimitris Galanis, John Pavlopoulos, Harris Papageorgiou, Ion
  Androutsopoulos, and Suresh Manandhar,
\newblock ``Semeval-2014 task 4: Aspect based sentiment analysis,''
\newblock in {\em Proc. of SemEval}, 2014.

\bibitem{unified_generative}
Hang Yan, Junqi Dai, Tuo Ji, Xipeng Qiu, and Zheng Zhang,
\newblock ``A unified generative framework for aspect-based sentiment
  analysis,''
\newblock in {\em Proc. of ACL}, 2021.

\bibitem{SentiPrompt}
Chengxi Li, Feiyu Gao, Jiajun Bu, Lu~Xu, Xiang Chen, Yu~Gu, Zirui Shao,
  Qi~Zheng, Ningyu Zhang, Yongpan Wang, and Zhi Yu,
\newblock ``Sentiprompt: Sentiment knowledge enhanced prompt-tuning for
  aspect-based sentiment analysis,''
\newblock {\em CoRR}, 2021.

\bibitem{generative1}
Yue Mao, Yi~Shen, Jingchao Yang, Xiaoying Zhu, and Longjun Cai,
\newblock ``Seq2path: Generating sentiment tuples as paths of a tree,''
\newblock in {\em Proc. of ACL Findings}, 2022.

\bibitem{neighboring}
Junqi Dai, Hang Yan, Tianxiang Sun, Pengfei Liu, and Xipeng Qiu,
\newblock ``Does syntax matter? {A} strong baseline for aspect-based sentiment
  analysis with roberta,''
\newblock in {\em Proc. of NAACL}, 2021.

\bibitem{neighboring2}
Kevin Clark, Urvashi Khandelwal, Omer Levy, and Christopher~D. Manning,
\newblock ``What does {BERT} look at? an analysis of bert's attention,''
\newblock in {\em Proceedings of the 2019 ACL Workshop BlackboxNLP: Analyzing
  and Interpreting Neural Networks for NLP, BlackboxNLP@ACL 2019, Florence,
  Italy, August 1, 2019}, 2019.

\bibitem{graph2}
Kai Sun, Richong Zhang, Samuel Mensah, Yongyi Mao, and Xudong Liu,
\newblock ``Aspect-level sentiment analysis via convolution over dependency
  tree,''
\newblock in {\em Proc. of EMNLP}, 2019.

\bibitem{graph3}
Kai Wang, Weizhou Shen, Yunyi Yang, Xiaojun Quan, and Rui Wang,
\newblock ``Relational graph attention network for aspect-based sentiment
  analysis,''
\newblock in {\em Proc. of ACL}, 2020.

\bibitem{graph4}
Hao Tang, Donghong Ji, Chenliang Li, and Qiji Zhou,
\newblock ``Dependency graph enhanced dual-transformer structure for
  aspect-based sentiment classification,''
\newblock in {\em Proc. of ACL}, 2020.

\bibitem{taki}
Taiqiang Wu, Xingyu Bai, Weigang Guo, Weijie Liu, Siheng Li, and Yujiu Yang,
\newblock ``Modeling fine-grained information via knowledge-aware hierarchical
  graph for zero-shot entity retrieval,''
\newblock {\em CoRR}, vol. abs/2211.10991, 2022.

\bibitem{graph1}
Bin Liang, Hang Su, Lin Gui, Erik Cambria, and Ruifeng Xu,
\newblock ``Aspect-based sentiment analysis via affective knowledge enhanced
  graph convolutional networks,''
\newblock {\em Knowl. Based Syst.}, 2022.

\bibitem{BART}
Mike Lewis, Yinhan Liu, Naman Goyal, Marjan Ghazvininejad, Abdelrahman Mohamed,
  Omer Levy, Veselin Stoyanov, and Luke Zettlemoyer,
\newblock ``{BART:} denoising sequence-to-sequence pre-training for natural
  language generation, translation, and comprehension,''
\newblock in {\em Proc. of ACL}, 2020.

\bibitem{GAT}
Petar Velickovic, Guillem Cucurull, Arantxa Casanova, Adriana Romero, Pietro
  Li{\`{o}}, and Yoshua Bengio,
\newblock ``Graph attention networks,''
\newblock {\em CoRR}, 2017.

\bibitem{penga}
Haiyun Peng, Lu~Xu, Lidong Bing, Fei Huang, Wei Lu, and Luo Si,
\newblock ``Knowing what, how and why: {A} near complete solution for
  aspect-based sentiment analysis,''
\newblock in {\em Proc. of AAAI}, 2020.

\bibitem{15res}
Maria Pontiki, Dimitris Galanis, Haris Papageorgiou, Suresh Manandhar, and Ion
  Androutsopoulos,
\newblock ``Semeval-2015 task 12: Aspect based sentiment analysis,''
\newblock in {\em Proc. of SemEval}, 2015.

\bibitem{16res}
Maria Pontiki, Dimitris Galanis, Haris Papageorgiou, Ion Androutsopoulos,
  Suresh Manandhar, Mohammad Al{-}Smadi, Mahmoud Al{-}Ayyoub, Yanyan Zhao, Bing
  Qin, Orph{\'{e}}e~De Clercq, V{\'{e}}ronique Hoste, Marianna Apidianaki,
  Xavier Tannier, Natalia~V. Loukachevitch, Evgeniy~V. Kotelnikov, N{\'{u}}ria
  Bel, Salud Mar{\'{\i}}a~Jim{\'{e}}nez Zafra, and G{\"{u}}lsen Eryigit,
\newblock ``Semeval-2016 task 5: Aspect based sentiment analysis,''
\newblock in {\em Proc. of SemEval}, 2016.

\bibitem{D17}
Wenya Wang, Sinno~Jialin Pan, Daniel Dahlmeier, and Xiaokui Xiao,
\newblock ``Coupled multi-layer attentions for co-extraction of aspect and
  opinion terms,''
\newblock in {\em Proc. of AAAI}, 2017.

\bibitem{D19}
Zhifang Fan, Zhen Wu, Xin{-}Yu Dai, Shujian Huang, and Jiajun Chen,
\newblock ``Target-oriented opinion words extraction with target-fused neural
  sequence labeling,''
\newblock in {\em Proc. of NAACL}, 2019.

\bibitem{RINANTE}
Hongliang Dai and Yangqiu Song,
\newblock ``Neural aspect and opinion term extraction with mined rules as weak
  supervision,''
\newblock in {\em Proc. of ACL}, 2019.

\bibitem{Li-unified}
Xin Li, Lidong Bing, Piji Li, and Wai Lam,
\newblock ``A unified model for opinion target extraction and target sentiment
  prediction,''
\newblock in {\em Proc. of AAAI}, 2019.

\bibitem{pengb}
Lu~Xu, Hao Li, Wei Lu, and Lidong Bing,
\newblock ``Position-aware tagging for aspect sentiment triplet extraction,''
\newblock in {\em Proc. of EMNLP}, 2020.

\bibitem{unified1}
Yue Mao, Yi~Shen, Chao Yu, and Longjun Cai,
\newblock ``A joint training dual-mrc framework for aspect based sentiment
  analysis,''
\newblock in {\em Proc. of AAAI}, 2021.

\bibitem{SynMux}
Hao Fei, Fei Li, Chenliang Li, Shengqiong Wu, Jingye Li, and Donghong Ji,
\newblock ``Inheriting the wisdom of predecessors: A multiplex cascade
  framework for unified aspect-based sentiment analysis,''
\newblock in {\em Proc. of IJCAI}, 2022.

\end{thebibliography}

\end{document}